\documentclass[10 pt, conference]{ieeeconf}
\usepackage{algorithm}
\usepackage{algpseudocode}
\usepackage{microtype}

\usepackage{subfig}
\usepackage{setspace}
\usepackage{amssymb, amsmath, times}
\usepackage{pifont}
\usepackage[export]{adjustbox}
\usepackage{footnote}
\usepackage{epstopdf, tikz}
\DeclareGraphicsExtensions{.eps}
\usepackage{schemabloc, verbatim, graphicx, array, booktabs}

\usepackage{soul}
\usepackage{breqn}
\usepackage{bm}
\usepackage{pgfplots}

\newcommand{\subparagraph}{}   
\usepackage[compact]{titlesec}
\titlespacing{\section}{0pt}{*0}{*0}
\titlespacing{\subsection}{0pt}{*}{*0}
\titlespacing{\subsubsection}{0pt}{*0}{*0}

\makeatletter
\let\NAT@parse\undefined
\makeatother

\usepackage{cite}
\usepackage{color}
\usepackage[pagebackref=true,breaklinks=true,letterpaper=true,colorlinks=true, citecolor=light-blue,linkcolor=cyan,urlcolor=blue,bookmarks=false]{hyperref}
\pgfplotsset{compat=1.5}
\definecolor{light-blue}{rgb}{0.30,0.35,1}
\definecolor{light-green}{rgb}{0.20,0.49,.85}

\newcounter{mnote}

\newcommand{\ie}{$i.e.$\ }
\newcommand{\eg}{e.g.\ }

\newcommand*\idx[2][]
{
\def\next{#1}%
\ifx\empty\next
  (#2)
\else
  (#1, #2)
\fi
}
\newcommand*\elt[3][]
{
\def\next{#1}%
\ifx\empty\next
  #2\idx{#3}
\else
  #1\idx{#2,#3}
\fi
}
\newcommand*\pd[3][]
{
\def\next{#1}%
\ifx\empty\next
  \frac{\partial#2}{\partial #3}
\else
  \frac{{\partial^{#1} #2}}{\partial#3^{#1}}
\fi
}

\newcommand{\mc}[1]{\mathcal{#1}}

\newcommand{\dist}{\mc{D}}

\def\nomdist{\bar{\textbf{v}}}
\def\control{\textbf{u}}
\def\admcontrol{\mathcal{U}}
\def\nomstate{\bar{\textbf{x}}}
\def\state{\textbf{x}}

\def\augreward{\ell}

\def\hinf{H_\infty}

\def\dist{\textbf{v}}
\def\distbo{\textbf{v}}

\def\tidx{t}
\def\fstate{f_{\state \tidx}}
\def\fcontrol{f_{\control \tidx}}
\def\fcontrolcontrol{f_{\control\control \tidx}}
\def\fcontroldistbo{f_{\control\distbo \tidx}}
\def\fdistbo{f_{\distbo \tidx}}
\def\fdistbodistbo{f_{\distbo \distbo \tidx}}
\def\fdistbostate{f_{\distbo \state \tidx}}
\def\fcontrolstate{f_{\control \state \tidx}}
\def\Vstate{V_{\state \tidx+1}}
\def\Vstateback{V_{\state \tidx}}
\def\Vstatestate{V_{\state \state \tidx+1}}
\def\Vstatestateback{V_{\state \state \tidx}}
\def\Qstate{Q_{\state \tidx}}
\def\Qcontrol{Q_{\control \tidx}}
\def\Qdistbo{Q_{\distbo \tidx}}
\def\Qstatestate{Q_{\state \state \tidx}}
\def\Qstatecontrol{Q_{\state \control \tidx}}
\def\Qcontrolstate{Q_{\control \state \tidx}}
\def\Qcontrolcontrol{Q_{\control \control \tidx}}
\def\Qstatecontrol{Q_{\state \control \tidx}}
\def\Qcontrolstate{Q_{\control \state \tidx}}
\def\Qcontrolcontrol{Q_{\control \control \tidx}}
\def\Qcontroldistbo{Q_{\control \distbo \tidx}}
\def\Qdistbostate{Q_{\distbo \state \tidx}}
\def\Qdistbodistbo{Q_{\distbo \distbo \tidx}}
\def\Qstatedistbo{Q_{\state \distbo \tidx}}
\def\Qdistbostate{Q_{\distbo \state \tidx}}
\def\Qdistbocontrol{Q_{\distbo \control \tidx}}

\def\TQcontrol{\tilde{Q}_{\control \tidx}}
\def\TQdistbo{\tilde{Q}_{\distbo \tidx}}

\def\TQcontrolstate{\tilde{Q}_{\control \state \tidx}}
\def\TQcontrolcontrol{\tilde{Q}_{\control \control \tidx}}

\def\TQcontrolstate{\tilde{Q}_{\control \state \tidx}}
\def\TQcontrolcontrol{\tilde{Q}_{\control \control \tidx}}
\def\TQcontroldistbo{\tilde{Q}_{\control \distbo \tidx}}
\def\TQdistbostate{\tilde{Q}_{\distbo \state \tidx}}
\def\TQdistbodistbo{\tilde{Q}_{\distbo \distbo \tidx}}

\def\TQdistbostate{\tilde{Q}_{\distbo \state \tidx}}
\def\TQdistbocontrol{\tilde{Q}_{\distbo \control \tidx}}

\def\cost{\mathcal{J}} 
\def\lstate{\ell_{\state \tidx}}
\def\lcontrol{\ell_{\control \tidx}}
\def\ldistbo{\ell_{\distbo \tidx}}
\def\lstatestate{\ell_{\state \state \tidx}}

\def\lcontrolstate{\ell_{\control \state \tidx}}
\def\lcontrolcontrol{\ell_{\control \control \tidx}}

\def\lcontrolstate{\ell_{\control \state \tidx}}
\def\lcontrolcontrol{\ell_{\control \control \tidx}}
\def\lcontroldistbo{\ell_{\control \distbo \tidx}}
\def\ldistbostate{\ell_{\distbo \state \tidx}}
\def\ldistbodistbo{\ell_{\distbo \distbo \tidx}}

\def\ldistbostate{\ell_{\distbo \state \tidx}}
\def\ldistbocontrol{\ell_{\distbo \control \tidx}}

\def\delstate{\delta \state}
\def\delcontrol{\delta \control}

\def\deldistbo{\delta \distbo}

\def\vectorgain{\textbf{\textit{g}}}
\def\matrixgain{\textbf{\textit{G}}}

\newcommand{\Note}[1]{}
\renewcommand{\Note}[1]{\hl{[#1]}}  

\newcolumntype{M}[1]{>{\centering\arraybackslash}m{#1}}
\def\coriolis{\textbf{\textit{C}}}
\def\massinertia{\textbf{\textit{M}}}
\def\torque{\bm{\tau}}
\def\frictionvec{\textbf{\textit{f}}}
\def\Smat{\textbf{\textit{S}}}
\def\Bmat{\textbf{\textit{B}}}
\def\wheelrad{\textbf{\textit{r}}}

\def\stateweight{\textbf{\textit{w}}_x}
\def\actionweight{\textbf{\textit{w}}_u}
\def\advactionweight{\textbf{\textit{w}}_v}

\title{\LARGE \textbf
Minimax Iterative Dynamic Game: Application to Nonlinear Robot Control Tasks.
	}

\author{Olalekan Ogunmolu, Nicholas Gans, and Tyler Summers.
	\thanks{Olalekan Ogunmolu and Nicholas Gans are with the Department of Electrical Engineering, Tyler Summers is with the Department of Mechanical Engineering, University of Texas at Dallas, Richardson, TX 75080, USA.
		{\tt\small \{olalekan.ogunmolu, ngans, tyler.summers\}@utdallas.edu}}%
}

\pdfoutput=1
\IEEEoverridecommandlockouts

\begin{document}
	\bstctlcite{IEEEexample:BSTcontrol}
	\maketitle
\textit{}	\thispagestyle{empty}
	\pagestyle{empty}

\begin{abstract}
Multistage decision policies provide useful control strategies in high-dimensional state spaces, particularly in complex control tasks.  However, they exhibit weak performance guarantees in the presence of disturbance, model mismatch, or model uncertainties. This brittleness limits their use in high-risk scenarios.
We present how to quantify the sensitivity of such policies  in order to inform of their robustness capacity. We also propose  a minimax iterative dynamic game framework for designing robust policies in the presence of disturbance/uncertainties.
We test the quantification hypothesis on a carefully designed deep neural network policy; we then pose a minimax iterative dynamic game (iDG) framework for improving policy robustness in the presence of adversarial disturbances. We evaluate our iDG framework on a mecanum-wheeled robot, whose goal is to find a ocally robust optimal multistage policy that achieve a given goal-reaching task.
The algorithm is simple and adaptable for designing meta-learning/deep policies that are robust against disturbances, model mismatch, or model uncertainties, up to a disturbance bound. Videos of the results are on the author's  \href{http://ecs.utdallas.edu/~opo140030/iros18/iros2018.html}{\underline{website}}, while the codes for reproducing our experiments are on \href{https://github.com/lakehanne/youbot/tree/rilqg}{\underline{github}}. A self-contained environment for reproducing our results is on \href{https://hub.docker.com/r/lakehanne/youbotbuntu14/}{\underline{docker}}. 
\end{abstract}

\section{Introduction}
Multistage decision policies are often brittle to deploy on real-world systems owing to their lack of robustness \cite{lin2017tactics, kos17} and the data inefficiency of the learning process.
%
%

Methods of designing scalable high-dimensional policies often rely on heuristics e.g. ~\cite{Pinto17}, which do not always produce repeatable results. Quite often, policies are learned under partial observability, but sampling with partial observations can be unstable ~\cite{garcia15}. In the presence of model uncertainties or model mismatch between the source and target environments~\cite{mandlekar17}, we must therefore devise policies that are robust to perturbations. Our goal in this paper is to provide an underpinning for designing robust policies, leveraging on methods from $\hinf$ control theory \cite{Zhou98}, dynamic programming (DP) \cite{Bellman1957}, differential dynamic programming (DDP) \cite{JacobsonMayne}, and iterative LQG ~\cite{Todorov04}. Essentially, we consider the performance of policies in the presence of various adversarial agents \cite{Basar99, Littman94, Morimoto05} using a minimax method.
%
%
While recent DRL techniques produce performance efficiency for agent tasks in the real world \cite{Mordatch15, Zhang16, Levine14, Levine16, montgomery2016guided, mnih2015human, mnih2016asynchronous}, there are sensitivity concerns that need to be addressed, \eg the trade-off between a system's nominal performance and its performance in the face of uncertainty or model mismatch. 

\noindent \textbf{Contributions}
 \begin{itemize}
 	\item We provide a framework for \textit{testing the brittleness of  a policy}:  Given a nominal policy, we pit a disturbing input against it, whose magnitude is controlled by a disturbance parameter, $\gamma$. If an adversarial agent causes significant performance degradation, then $\gamma$ indicates the upper bound on the efficiency of such policy.
 	\item In an \textit{iterative, dynamic two player zero-sum Markov game} (iDG), we let each agent execute an opposite reaction to its pair: a concave-convex problem ensues, and we seek to drive the policy toward a saddle point equilibrium. 
 	 This iDG framework  generates  local control laws -- providing an alternating best response update of global control and adversarial policies, leading to a saddle-point equilibrium convergence. This is essentially a meta-algorithm that can be extended to quantify and design the robustness of model-free, model-based RL, as well as DP/iterative LQG family of policies.
 \end{itemize}

%
We evaluate our proposal for control of robot motor tasks using policy search \cite{PetersSurvey, Levine14} and the ILQG algorithm \cite{Todorov04}. The rest of this paper is thus organized:
we provide a formal treatment policy sensitivity quantification and the iDG algorithm within a linearly solvable MDP \cite{linearMDP} in \autoref{sec:two_player}. The dynamics and model of the robot we that evaluates our iDG hypothesis is presented in \autoref{sec:youbot_dyn}. Results for the two proposals of this work are discussed in \autoref{sec:results}, and we conclude this work in \autoref{sec:conclusions}. A more extensive discussion of the modeling methods in this paper is detailed in ~\cite{arxiv_article}.

\section{Two-Player Trajectory Optimization}   \label{sec:two_player}
Consider two agents interacting in an environment, $\mc{E}$, over a finite horizon, $T$; the states evolve according to a discrete-time stochastic dynamics,
\[
\state_{t+1} = f_t(\state_t, \control_t, \distbo_t), \, \, t = 0, \ldots, T-1, \,\,
\state_0 = \bar \state_0,
\]
where $\state_t \subseteq \mathcal{X}_t$ is the $n-$dimensional state vector, $\control_t \subseteq \admcontrol_t$ is the $m$-dimensional nominal agent's action (or control law), and $\distbo_t \in \mathcal{V}_t$ denote the disturbing agent's $p$-dimensional action. 
The nominal agent chooses its action under a (stochastic) policy $\{\pi = \pi_0, \pi_1, \ldots, \pi_{T}\} \subseteq \Pi$, while the uncertainty's actions are governed by a policy $\{\psi = \psi_0, \psi_1, \ldots , \psi_{T}\} \subseteq \Psi$. For the policy pair($\pi, \psi$), we define the \textit{cost-to-go}, $\cost(\state, \pi, \psi)$, of a trajectory $\{\state_t\}_{t = 0, \ldots, T}$ with initial condition $\state_0$,  as a partial sum of costs from $t$ to $T$,
\[
\cost_0(\state_0, \pi, \psi) = \mathbf{E}_{\control_t, \dist_t}  \sum_{t=0}^{T-1}  \ell_t (\state_t, \control_t, \dist_t) +
L_T(\state_T),
\]
where $\ell_t$ is a nonnegative function of $(\state_t, \control_t, \dist_t)$, denoting the stage cost, and $L_T$ is a nonnegative function of $\state_T$, denoting the final cost. We seek a pair of \emph{saddle point equilibrium} policies, $(\pi^*, \psi^*)$ that satisfy,
\[
\cost_0(\state_0, \pi^*, \psi) \leq\cost_0(\state_0, \pi^*, \psi^*) \leq \cost_0(\state_0, \pi, \psi^*),
\]
$ \forall \, \pi \in \Pi, \psi \in \Psi$ and $\state_0$.
For the general case where we start from an initial condition $\state_t$, one may write the dynamic programming (DP) equation above as
\[
\mc{J}_t^*(\state_t) = \min_{\pi \in \Pi} \max_{\psi \in \Psi} \mc{J}_t(\state_t, \pi, \psi),
\]
where $\pi$ and $\psi$ contain the control sequences $\{\control_t\}$ and $\{\dist_t\}$, and $\mc{J}(\cdot)$ denote the Hamilton-Jacobi Bellman cost function. The saddle point equilibrium for an optimal control sequence pair $\{\control_t^*, \dist_t^*\}$  can be obtained with
\begin{align}
\cost_t^*(\state_t) &\mbox{=} \min_{\pi \in \Pi} \max_{\psi \in \Psi} \cost_t(\state_t, \pi, \psi) 
\label{eq:minimax} \\
&\mbox{=} \min_{\pi \in \Pi} \max_{\psi \in \Psi} [\ell_t(\state_t, \control_t, \distbo_t) + \cost^*_{t+1}\left(f_t\left(\state_t, \control_t, \dist_t\right)\right)] \nonumber 
\end{align}
%
%
\subsection{Quantifying a policy's robustness}
Suppose that the nominal policy, $\pi$, of an agent has been found, and consider an interacting disturbuing agent so that the closed-loop dynamics is describable by the discretized Euler equation,
\begin{align}
\begin{aligned}
\state_{t+1} &= f_t( \state_t, \control_t,  \dist_t), \quad \control_t \sim \pi_t \\
&= \bar f_t(\state_t, \dist_t), \quad t = 0, ..., T-1.
\end{aligned}
\label{eq:nlnr_cl}
\end{align}
For stage costs of the form, $\ell_t(\state_t, \control_t, \distbo_t) = c_t(\state_t, \control_t) - \gamma g_t(\dist_t)$,
where $c_t(\state_t, \control_t)$ represents the nominal stage cost, $g_t(\cdot)$ is a norm on the adversarial input\footnote{This formulation takes $\mathcal{V}_t$ as a vector space but one can as well define a nonnegative adversarial penalty term when $\mathcal{V}_t$ is a finite set.}, penalizing the adversary's actions, and $\gamma > 0$ is a disturbance term that controls the strength of the adversary; the adversary faces a maximization problem of the form
\begin{align}
&\max_{\psi \in \Psi} \, \mathbf{E}_{\control_t \sim \pi_t}  \sum_{t=0}^{T}  c(\state_t, \control_t) - \gamma g(\distbo_t) = \max_{\psi \in \Psi} \, \mathbf{E} \sum_{t=0}^{T}  \bar \ell_t^\gamma(\state_t, \distbo_t). \nonumber
\end{align}
%
Varying $\gamma$ increases/decreases the penalty incurred by the adversarial agent's actions. As $\gamma \rightarrow \infty$, the adversary's optimal policy is to do nothing, since any action will incur an infinite penalty; as $\gamma$ decreases, the adversary incurs lower penalties, causing large system disturbance. The (inverse of the) smallest $\gamma$-value for which the adversary causes unacceptable performance degradation (e.g., instability) provides a measure of robustness of the nominal agent's policy $\pi$.
The parameter $\gamma$ is a distinguishing feature of our work; $\gamma$ quantifies the $\mathcal{H}_\infty$ norm of the closed-loop system, a measure of its robustness to an adversarial input. The adversary need not represent a malicious input, but can be interpreted as a worst possible disturbance of a given magnitude.
\begin{figure}[tb]
	\centering
	\includegraphics[width=.95\columnwidth,height=.65\columnwidth]{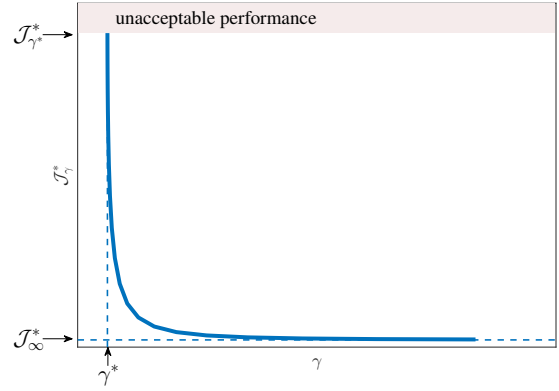}
	\caption{Prototypical Performance-Robustness Tradeoff Curve. 
	}
	\label{fig:tradeoff}
\end{figure}
%

In systems with nonlinear dynamics, one can use differential dynamic programming (DDP) \cite{JacobsonMayne} or iterative LQG \cite{Tassa12} to optimize the \emph{adversary} against the closed-loop system under the nominal agent's policy $\pi$. Indeed, Morimoto (in \cite{Morimoto03}) applied minimax DDP to a walking biped robot, but we noticed errors in the value function recursions\footnote{These corrections are given in \eqref{eq:value_coeffs}.}. We are motivated by the improved convergence guarantees of iterative LQG (iLQG) over DDP, and its suitability for solving constrained nonlinear control problems by iteratively linearizing a nonlinear system about a local neighboring trajectory and computing the optimal local control laws.
Our minimax iLQG framework facilitates learning control decision strategies that are robust in the presence of disturbances and modeling errors -- improving upon nominal iLQG policies.

A performance-robustness tradeoff curve can be computed by optimizing the adversary for various values of $\gamma$. A prototypical curve is illustrated in ~\autoref{fig:tradeoff}. This curve quantifies the robustness of a fixed policy $\pi$, defining an upper bound $\mc{J}_{\gamma^*}^*$ on the cost function that  does not yield acceptable performance: this is the critical value of $\gamma$ to be identified. This $\gamma^*$-value determines how much adversarial effort we need to suffer performance degradation; large $\gamma^*$ values will give poor robustness of $\pi$ -- even with little adversarial effort.
The algorithm for optimizing adversaries and generating the trade-off curve is summarized in Algorithm \ref{alg:robustness}.
The adversary's policy may not be globally optimal due to the approximation of policies in continuous spaces and because the algorithm may only converge to a locally optimal solution, we obtain an upper bound on the robustness of the policy $\pi$.
\begin{algorithm}[tb!]
	\caption{Robustness Curve via iLQG
		\label{alg:robustness}}
	\begin{algorithmic}[1]
		\For{$\gamma \in \text{range}(\gamma_{min}, \gamma_{max}, N)$}
		\For{$i \in \{1, \dots, I\}$}
		\State Initialize $x_0 = x(0)$
		\State  $p_i \leftarrow \arg\min_{p_i} \mathbf{E}_{p_i}\left[\sum_{t=0}^T \tilde{ \bar \ell}_t^\gamma(\state_t, \distbo_t) \right]$ 
		\EndFor
		\EndFor
		\label{step:ilqg}
	\end{algorithmic}
\end{algorithm}

\subsection{Achieving Robustness via IDG}
We propose to arrive at a saddle point equilibrium with \eqref{eq:minimax} by continually solving an online finite-horizon trajectory optimization problem. 
This is essentially a minimax framework and is applicable  in principle to any off/on-policy RL algorithm.
In this work, we generalize the online trajectory optimization algorithm of \cite{Tassa12}, to a two-player, zero-sum dynamic game as follows:
\begin{itemize}
\item we  approximate the nonlinear system dynamics (c.f. \eqref{eq:nlnr_cl}), starting with a schedule of the nominal agent's local controls, $\{ \bar \control_t\}$, and nominal adversarial agent's local controls $\{ \bar \distbo_t \}$ which are assumed to be available (when these are not available, we can initialize them to $0$)
\item  we then run the system's passive dynamics with $\{\bar{\control}\}, \{\bar {\dist} \}$ to generate  a nominal state trajectory $ \{\bar{\state}_t\}$, with neighboring trajectories $\{ \state_t \}$ 
\item we choose a small neighborhood, $\{\delstate_t\}$ of $\{\state_t\}$, which provides an optimal reduction in cost as the dynamics no longer represent those of $\{\state_t\}$
\item  discretizing time, the new state and control sequence pairs become $\delta \state_\tidx = \state_\tidx - \bar \state_\tidx, \,
\delta \control_\tidx = \control_\tidx - \bar \control_\tidx, \,
\delta \distbo_\tidx = \distbo_\tidx - \bar \distbo_\tidx.$
\end{itemize}
Setting $V(\state, T) = \augreward_{T}(\state_{T})$, the min-max over the entire control sequence reduces to a stepwise optimization over a single control, going backward in time with 
\begin{align}
V(\state_\tidx) = \min_{\control_\tidx \sim \pi} \max_{\distbo_\tidx \sim \psi} [\augreward(\state_\tidx, \control_\tidx, \distbo_\tidx) + V(f(\state_{\tidx+1}, \control_{\tidx+1}, \distbo_{\tidx+1}))]. \nonumber
\end{align}
%
If we consider the Hamiltonian, $\augreward(\cdot)+V(\cdot)$, as a perturbation around the tuple $\{\state_\tidx, \control_\tidx, \distbo_\tidx\}$, the cost over the local neighborhood via an approximate Taylor series expansion becomes
\[
Q_\tidx = \augreward(\state_\tidx, \control_\tidx, \distbo_\tidx, \tidx) + V(\state_{\tidx+1}, \tidx+1).
\]
A second-order approximation of the perturbed $Q$-coefficients of the LQR problem around the neighborhood $\{ \delstate_\tidx \}$ of the trajectory $\{\state_\tidx\}$ is defined as,
\begin{equation} 	\label{eq:taylor_expand}
\small
Q(\cdot)
\approx
\dfrac{1}{2}
\begin{bmatrix}
1 \\ \delstate_\tidx^T \\ \delcontrol_\tidx^T \\ \deldistbo_\tidx^T
\end{bmatrix}^T
%
\begin{bmatrix}
1 & \Qstate^T & \Qcontrol^T & \Qdistbo^T \\
\Qstate & \Qstatestate & \Qstatecontrol & \Qstatedistbo \\
\Qcontrol & \Qcontrolstate & \Qcontrolcontrol & \Qcontroldistbo	\\
\Qdistbo & \Qdistbostate & \Qdistbocontrol & \Qdistbodistbo
\end{bmatrix}
\begin{bmatrix}
1 \\ \delstate_\tidx \\ \delcontrol_\tidx \\ \deldistbo_\tidx
\end{bmatrix},
\end{equation}
where,
\begin{equation}
\label{eq:Q_coeffs}
\small
\begin{aligned}
\Qstate  &= \augreward_{\state \tidx} + \fstate^T \Vstate, \quad \Qcontrol = \augreward_{\control \tidx} + \fcontrol^T \Vstate  \nonumber \\ 
\Qdistbo &= \augreward_{\distbo \tidx} + \fdistbo^T \Vstate, \quad \Qstatestate = \augreward_{\state \state \tidx} + \fstate^T \Vstatestate \fstate  \\
\Qcontrolstate  &= \augreward_{\control \state \tidx} + \fcontrol^T \Vstatestate \fstate, \,
\Qdistbostate = \augreward_{\distbo \state \tidx} + \fdistbo^T \Vstatestate \fstate  \nonumber \\
\Qcontrolcontrol &= \augreward_{\control \control \tidx} + \fcontrol^T \Vstatestate \fcontrol, \, \Qdistbodistbo  = \augreward_{\distbo \distbo \tidx} + \fdistbo^T \Vstatestate \fdistbo \nonumber \\
\,
\Qcontroldistbo &= \lcontroldistbo + \fcontrol^T \Vstatestate \fdistbo. \nonumber
\end{aligned}
\end{equation}
This is consistent with linearized methods, where linearized second moment terms will dominate the higher order terms.
The LQR approximation to the state and the optimal control performance index become,
\begin{align}
\begin{split}
&\delstate_{\tidx + 1} \approx \fstate \delstate_\tidx + \fcontrol \delcontrol_\tidx + \fdistbo
\deldistbo_\tidx 
\label{eq:dyn_cost}
 \\
\small
\augreward(\state_\tidx, \control_\tidx, \dist_\tidx) \approx
\small
\dfrac{1}{2}
&
\small
\begin{bmatrix}
1 \\ \delstate_\tidx^T \\ \delcontrol_\tidx^T \\ \deldistbo_\tidx^T
\end{bmatrix}^T
\small 
\begin{bmatrix}
\ell_{0t} & \lstate^T & \lcontrol^T & \ldistbo^T \\
\lstate & \lstatestate & \lcontrolstate^T & \ldistbostate^T \\
\lcontrol & \lcontrolstate & \lcontrolcontrol & \lcontroldistbo	\\
\ldistbo & \ldistbostate & \ldistbocontrol & \ldistbodistbo
\end{bmatrix}
\small
\begin{bmatrix}
1 \\ \delstate_\tidx \\ \delcontrol_\tidx \\ \deldistbo_\tidx
\end{bmatrix}
\end{split}
\end{align}
where single and double subscripts denote first and second-order derivatives\footnote{ Note that $\delta \state_k, \delta \control_k, \text{ and } \delta \distbo_k$ are measured w.r.t the nominal vectors $\bar \state_k, \bar \control_k, \bar \distbo_k$ and are not necessarily small.}. 
%
The best possible nominal agent's action and the worst possible adversarial action can be found by performing the respective arg min and arg max operations on the $Q-$ function in \eqref{eq:taylor_expand} so that
\begin{align} \label{eq:local_controls}
\begin{split}
\delcontrol_\tidx^\star  &= -\Qcontrolcontrol^{-1}\left[\Qcontrol^T + \Qcontrolstate \delstate_\tidx + \Qcontroldistbo \deldistbo_\tidx \right],
\\
\deldistbo_\tidx^\star  &= -\Qdistbodistbo^{-1}\left[\Qdistbo^T + \Qdistbostate \delstate_\tidx + \Qdistbocontrol \delcontrol_\tidx \right].
\end{split}
\end{align}
Note that the control strategies in \eqref{eq:local_controls} depend on the action of the other player. Say the nominal agent first implements its strategy, then transmits its information to the adversary, which subsequently chooses its strategy; it follows that the adversary can choose a more favorable outcome since it knows what the nominal agent's strategy is. It becomes obvious that the \textit{best} action for the nominal agent is to choose a control strategy that is an optimal response to the action choice of the adversary. Similarly, if the roles of the players are changed, the nominal agent's response to the adversary's \textit{worst} choice will be more favorable since it knows what the adversarial agent's strategy is. Therefore, it does not matter that the order of play is predetermined. We end up with an  \textit{iterative dynamic game}, where each agent's strategy depends on its opponent's actions.
This ensures that we have a \textit{cooperative game} in which the nominal and adversarial agent alternate between taking best possible and worst possible actions during the trajectory optimization phase. This helps maintain equilibrium around the system's desired trajectory, while ensuring robustness in local policies. Suppose we set,
\begin{align}
\textbf{K}_{\control \tidx} &=  \left[\left(I - \Qcontrolcontrol^{-1}\Qcontroldistbo \Qdistbodistbo^{-1}\Qcontroldistbo^T\right)\Qcontrolcontrol^{-1}\right]^{-1}, \nonumber \\
\textbf{K}_{\distbo \tidx} &= \left[\left(I - \Qdistbodistbo^{-1}\Qcontroldistbo^T \Qcontrolcontrol^{-1}\Qcontroldistbo\right)\Qdistbodistbo^{-1}\right]^{-1}, 
\nonumber \\
%
\vectorgain_{\control \tidx} &= \textbf{K}_{\control \tidx} (\Qcontroldistbo \Qdistbodistbo^{-1}\Qdistbo - \Qcontrol),
 \nonumber \\
\vectorgain_{\distbo \tidx}  &= \textbf{K}_{\distbo \tidx} (\Qcontroldistbo^T \Qcontrolcontrol^{-1}\Qcontrol - \Qdistbo),
\nonumber \\
\matrixgain_{\control \tidx} &=  \textbf{K}_{\control \tidx} \left( \Qcontroldistbo \Qdistbodistbo ^{-1}\Qdistbostate - \Qcontrolstate\right),
\nonumber \\
\matrixgain_{\distbo \tidx} &= \textbf{K}_{\distbo \tidx} \left( \Qcontroldistbo^T \Qcontrolcontrol ^{-1}\Qcontrolstate - \Qdistbostate\right),  \nonumber
\end{align}
then it follows that we can rewrite \eqref{eq:local_controls} as
\begin{align}  \label{eq:prot_and_adv}
\delcontrol^\star_\tidx = \vectorgain_{\control_\tidx} + \matrixgain_{\control_\tidx} \delstate_\tidx, \quad \deldistbo^\star_\tidx = \vectorgain_{\distbo_\tidx} + \matrixgain_{\distbo_\tidx} \delstate_\tidx.
\end{align}
Equation \ref{eq:prot_and_adv} gives the open and closed-loop components of the control equations for both agents.
Comparing coefficients in \eqref{eq:Q_coeffs}, we find that the value function coeeficients can be thus written
\begin{align}  
&\Delta V_\tidx = \vectorgain_{\control \tidx} \Qcontrol
+  \vectorgain_{\distbo \tidx} \Qdistbo
+  \vectorgain_{\control \tidx} \Qcontroldistbo \vectorgain_{\distbo \tidx}
 \nonumber \\
&\quad +\frac{1}{2} \left(\vectorgain_{\control \tidx} \Qcontrolcontrol \vectorgain_{\control \tidx}
+ \vectorgain_{\distbo \tidx} \Qdistbodistbo \vectorgain_{\distbo \tidx} \right)
\nonumber \\
\Vstateback =\Qstate
+ \matrixgain_{\control_\tidx}^T&\Qcontrol
+  \matrixgain_{\distbo_\tidx}^T \Qdistbo
+ \matrixgain_{\control \tidx}^T \Qcontrolcontrol \vectorgain_{\control \tidx}
+ \vectorgain_{\control \tidx} \Qcontrolstate
\nonumber \\
+ \vectorgain_{\distbo \tidx} \Qdistbostate
+  \matrixgain_{\distbo \tidx}^T &\Qdistbodistbo \vectorgain_{\distbo \tidx} 
%
+ \matrixgain_{\distbo \tidx}^T \Qcontroldistbo^T \vectorgain_{\control \tidx}
+ \matrixgain_{\control \tidx}^T \Qcontroldistbo \vectorgain_{\distbo \tidx}
\nonumber \\
\Vstatestateback = \frac{1}{2}(\Qstatestate
+ &\matrixgain_{\control_\tidx}^T \Qcontrolcontrol \matrixgain_{\control_\tidx}
+ \matrixgain_{\distbo_\tidx}^T \Qdistbodistbo \matrixgain_{\distbo_\tidx}) +  \matrixgain_{\control_\tidx}^T \Qcontrolstate  \nonumber \\
&\qquad
+   \matrixgain_{\distbo_\tidx}^T \Qdistbostate
+  \matrixgain_{\control_\tidx}^T \Qcontroldistbo \matrixgain_{\distbo_\tidx}.
\label{eq:value_coeffs}
\end{align}
\textit{These recursive value functions, essentially 
differentiate our value coefficient recursion equations from Morimoto's DDP recursions}.
%
\subsection{Improved Regularization}
For nonlinear systems, the inverse of the Hessian must be strictly positive definite. When the inverse of the Hessian is non-positive-definite, we can add a suitably large positive quantity to it \cite{Kelley63, BullockFranklin67}, or replace the elements of the diagonal matrix in its eigen decomposition, $[V, \mathcal{D}] = eig(Q)$, that are smaller than an adequately small $\rho$, and then set $Q = V \mathcal{D} V^T$ \cite{Todorov04}. In this work, $\rho$ is added to $\textbf{Q}$ when the Hessian is not well-posed. Our update rule is $\tilde{\textbf{Q}}_{\control\control} = \textbf{Q}_{\control\control} + \rho\textbf{I}_m.$
%
However, the regularization can have adverse effects on the system arising from the control/disturbance transition matrices $\fcontrol$ and $\fdistbo$; therefore, we introduce a similar penalty term used in \cite{Tassa12} to deviations from states so that the regularization yields a quadratic state cost about the previous policy:
\begin{align}
		\tilde{\textbf{Q}}_{\control\control\tidx} &= \lcontrolcontrol + \fcontrol^T (\Vstatestate + \rho \textbf{I}_n)\fstate + \Vstate \fcontrolcontrol \nonumber \\
		\tilde{\textbf{Q}}_{\distbo\distbo\tidx} &= \ldistbodistbo + \fdistbo^T (\Vstatestate + \rho \textbf{I}_n)\fstate + \Vstate \fdistbodistbo \nonumber \\
		\tilde{\textbf{Q}}_{\control\state\tidx} &= \lcontrolstate +  \fcontrol^T (\Vstatestate + \rho \textbf{I}_n)\fstate  + \Vstate \fcontrolstate \nonumber \\
		\tilde{\textbf{Q}}_{\distbo\state\tidx} &= \ldistbostate +  \fdistbo^T (\Vstatestate + \rho \textbf{I}_n)\fstate + \Vstate \fdistbostate.
\end{align}
The adjusted gains therefore become
\begin{align}
\tilde{\textbf{K}}_{\control \tidx} &=  \left[\left(I - \TQcontrolcontrol^{-1}\TQcontroldistbo \TQdistbodistbo^{-1}\TQcontroldistbo^T\right)\TQcontrolcontrol^{-1}\right]^{-1}, \nonumber \\
\tilde{\textbf{K}}_{\distbo \tidx} &= \left[\left(I - \TQdistbodistbo^{-1}\TQdistbocontrol \TQcontrolcontrol^{-1}\TQcontroldistbo\right)\TQdistbodistbo^{-1}\right]^{-1}
\nonumber \\
\vectorgain_{\control \tidx} &= \tilde{\textbf{K}}_{\control \tidx} (\TQcontroldistbo \TQdistbodistbo^{-1}\TQdistbo - \TQcontrol),
\nonumber \\
\matrixgain_{\control \tidx} &=  \tilde{\textbf{K}}_{\control \tidx} \left( \TQcontroldistbo \TQdistbodistbo ^{-1}\TQdistbostate - \TQcontrolstate\right) \nonumber \\
\vectorgain_{\distbo \tidx}  &= \tilde{\textbf{K}}_{\distbo \tidx} (\TQdistbocontrol \TQcontrolcontrol^{-1}\TQcontrol - \TQdistbo),
\nonumber \\
\matrixgain_{\distbo \tidx} &= \tilde{\textbf{K}}_{\distbo \tidx} \left( \TQdistbocontrol \TQcontrolcontrol ^{-1}\TQcontrolstate - \TQdistbostate\right).
\end{align}
\noindent The improved value functions are updated in \eqref{eq:value_coeffs} accordingly.
\subsection{Regularization Schedule}
To accurately tweak the regularizarion term, $\rho$, we adopt a regularization schedule that penalizes $\Qcontrolcontrol$ or $\Qdistbodistbo$ when the backward pass fails. When the backward pass is successful, we would desire rapid decrease in $\rho$  in order to assure fast convergence; otherwise, we would want to quickly increase $\rho$, albeit in a bumpless manner since the minimum value of $\rho$ that prevents divergence is of linear order. We let $\rho_0$ denote some minimal modification value for $\rho$ (set to $1.0$), and we adjust $\rho$ as follows:
\[
\begin{array}{cc}
\begin{tabular}{l}
\textbf{increase} $\rho$:  \\
$\rho \leftarrow 1.1 \, \rho_0$  \\
$\rho_0 = \rho$  
\end{tabular}&
\begin{tabular}{l} 
\textbf{reduce} $\rho$:  \\
$\rho \leftarrow 0.09 \, \rho_0$ \\
$\rho_0 = \rho$
\end{tabular}
\end{array}
\]

\section{Dynamics Modeling and Simulation}   \label{sec:youbot_dyn}
 \begin{figure}[tb]
	\centering
	\subfloat[Mecanum Wheels Model]
	{\includegraphics[trim={1in 0in 1in 0in},width=0.5\columnwidth,clip]{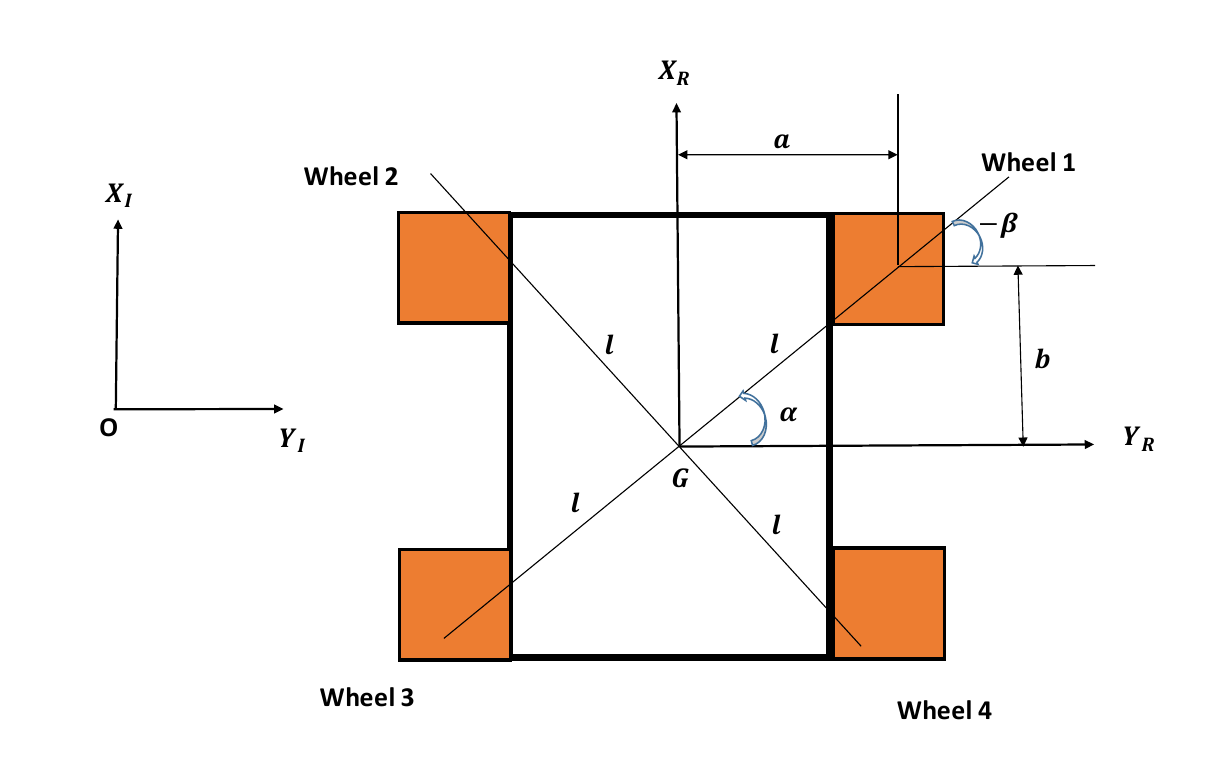}\label{fig:mecanum}}
	~ 
	\subfloat[Robot frames convention	]
	{\includegraphics[trim={1in 0in 1in 0in},width=0.5\columnwidth,clip]{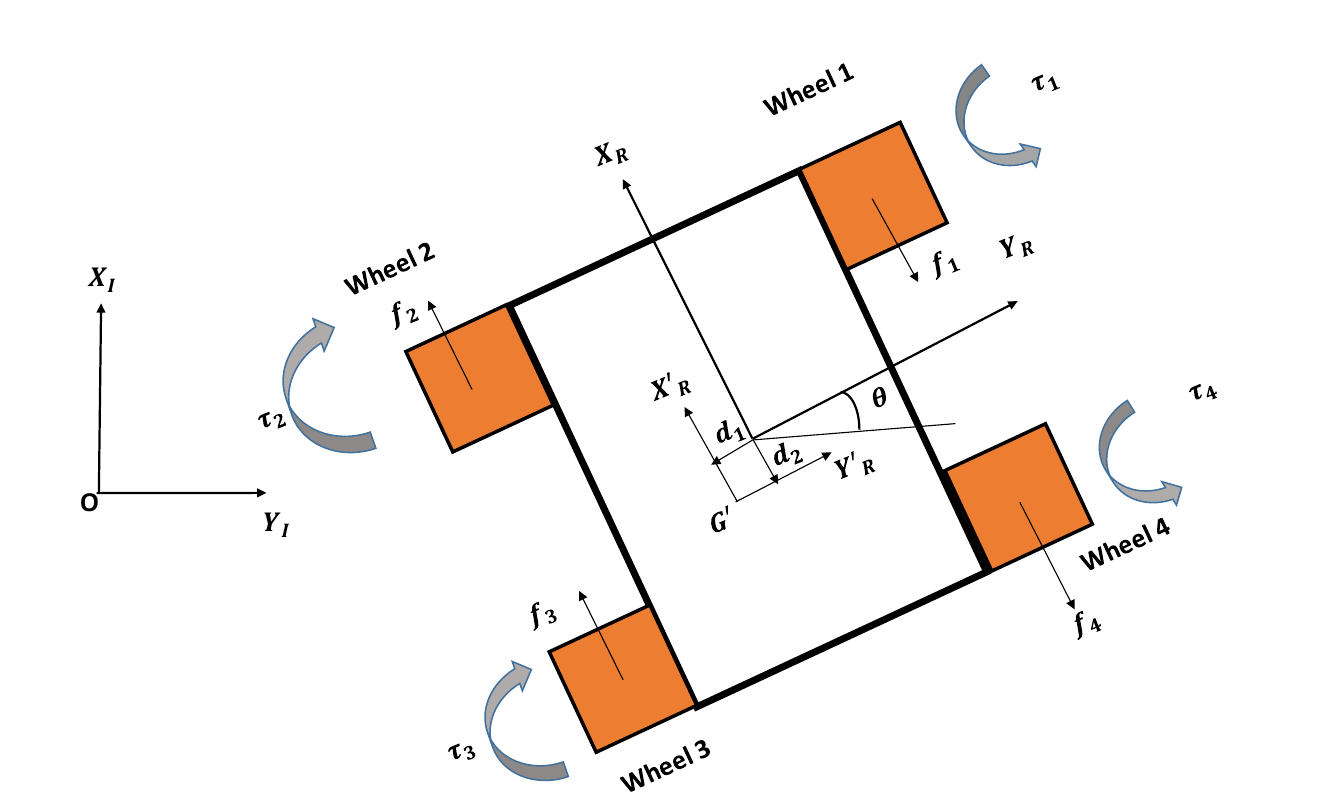}\label{fig:robot_geom}}
	\caption{Robot Geometry}
\end{figure}
We consider the KUKA youbot\footnote{\href{https://goo.gl/CYTjvD}{https://goo.gl/CYTjvD}} platform with four mecanum wheels, capable of spatial $\{x,y\}$ motion, \ie sideways, and forward, and an in-place $\theta$-rotation about the $z$-axis (see Fig. \autoref{fig:mecanum}). It is equipped with a 5-DoF arm, mounted on its base. We use the complete kinematic and dynamic model of the youbot platform, accounting for the wheels' friction and mass, while neglecting the links' masses and their associated inertia forces. 
The coordinates of the robot in the world frame are  denoted $\textit{\state}_R = \begin{bmatrix}x_R, y_R, \theta_R\end{bmatrix}^T$, where given as the $x_R,\,y_R$ are coordinates of the origin of the robot frame and $\theta_R$ is the relative angle between the  world and robot $x$ axes (see Fig. \autoref{fig:robot_geom}).
%
%

The torques that govern the robot's motion are obtained from \cite{mecanum}. We run our experiments in the Gazebo physics engine, which has its reference frame defined as $x$ pointing forward, $y$ pointing sideways, and $z$ pointing up. Therefore, our reference frame and robot geometry are as illustrated in Figs \ref{fig:mecanum} and \ref{fig:robot_geom}. Formally, we define the generalized Lagrangian equation of the robot as,
\begin{align}
\massinertia(\textit{\state})\ddot{\textit{\state}} + \coriolis(\textit{\state}, \ddot{\textit{\state}}) \dot{\textit{\state}} + \Bmat^T \Smat \frictionvec = \frac{1}{\wheelrad}\Bmat^T \torque
\label{eq:inverse_dyn}
\end{align}
where $\torque = [\torque_1, \torque_2, \torque_3, \torque_4]$ is the wheel torque vector, $\wheelrad$ is the wheel radius, $\frictionvec = [\frictionvec_1, \frictionvec_2, \frictionvec_3, \frictionvec_4]^T$ is the friction vector, and $\Smat$ and $\Bmat$ map the inverse kinematics, gravity, external forces and robot's angle, $\theta$, 
to each wheel torque; matrices $\massinertia$ and $\coriolis$ denote the inertia and coriolis properties of the robot. $\Bmat$ and $\Smat$ are given by,
\[
\small
	\Bmat =
	\begin{bmatrix}
	-(\cos \theta - \sin \theta) & -(\cos \theta + \sin \theta) & -\sqrt{2}l \sin(\zeta) \\
	-(\cos \theta + \sin \theta) & (\cos \theta - \sin \theta) &- \sqrt{2}l \sin(\zeta) \\
	(\cos \theta - \sin \theta) & (\cos \theta + \sin \theta)  & -\sqrt{2}l \sin(\zeta) \\
	(\cos \theta + \sin \theta) & -(\cos \theta - \sin \theta)  & -\sqrt{2}l \sin(\zeta)
	\end{bmatrix}
\]
\[
\Smat = \text{diag} \begin{bmatrix}
	sgn(\dot{\phi}_1), \,  sgn(\dot{\phi}_2),  \, sgn(\dot{\phi}_3), \,  sgn(\dot{\phi}_4);
\end{bmatrix}
\]
$\zeta=\pi/4 - \alpha$, $l$ is the mounting distance of the wheels as shown in  Fig. \autoref{fig:mecanum}, and
$\dot{\phi}_i$, is the rotation speed of each wheel about its axis of rotation. We apply the generalized force/torque vector, $F_{i}$, to the base frame of the robot, defined as, 
\begin{align}
	F_{i} = \sum_{j=1}^{4}\left(\torque_j - r \, sgn(\dot{\phi_j}) \, \frictionvec_j\right)\frac{\partial \dot{\phi}_j}{\partial \dot{\textit{\state}}_i}, \, i = \{1,2,3\}
\end{align}

\begin{figure}[tb!]
	\centering
	\subfloat[Home Position.
	]{		\includegraphics[height=0.35\columnwidth,width=0.45\columnwidth]{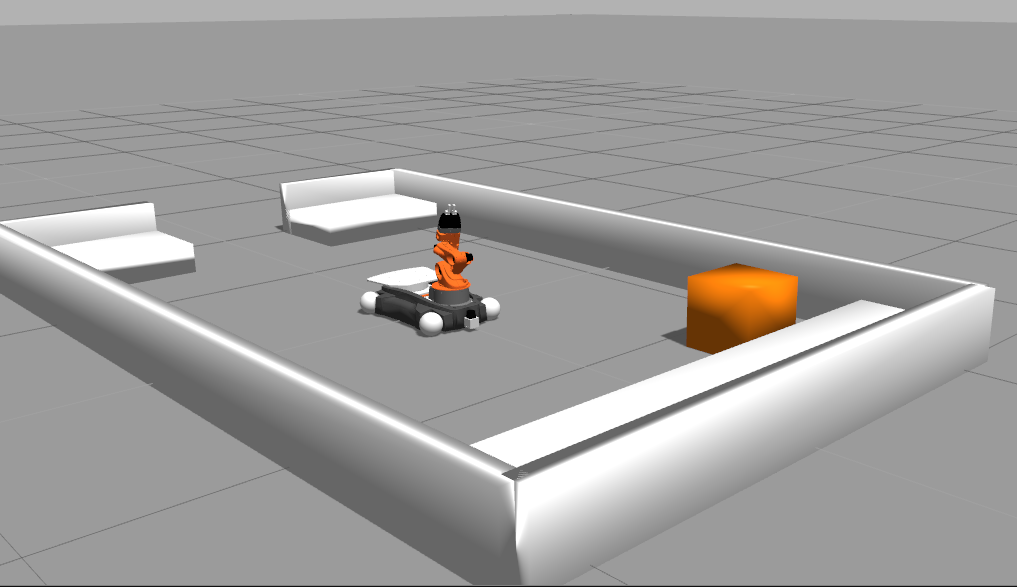}}%
	~ 
	\subfloat[Goal State.
	]{\includegraphics[height=0.35\columnwidth,width=0.45\columnwidth]{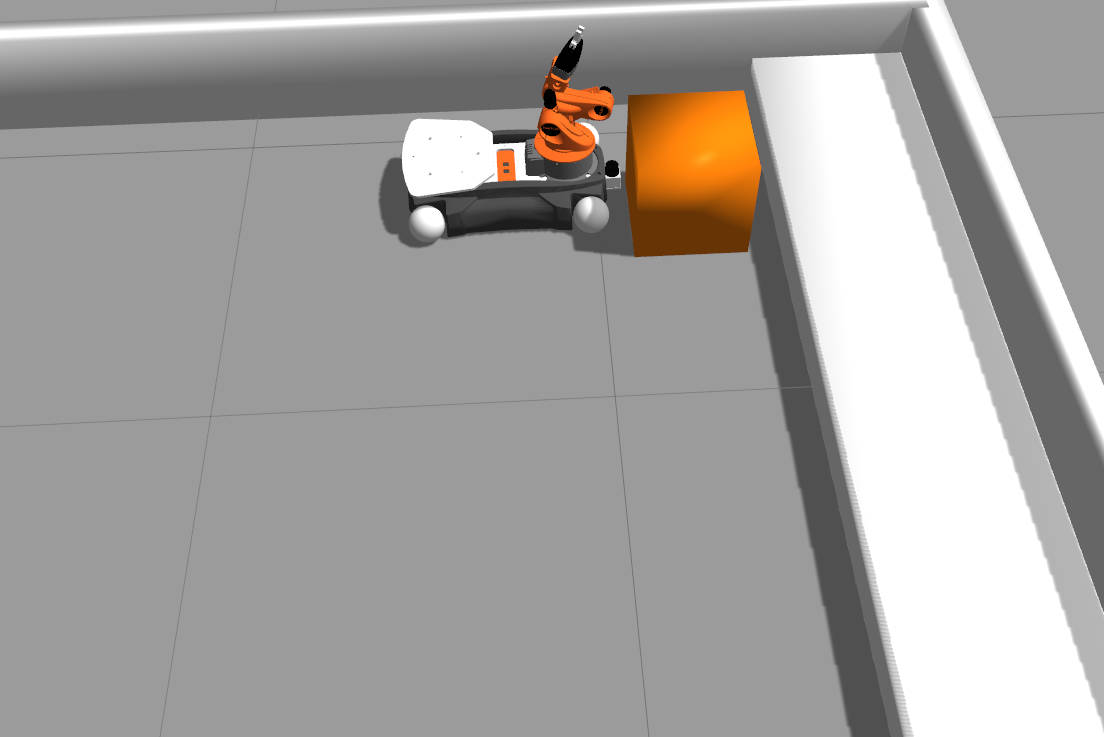}}%
	\caption{Goal Navigation Illustration}
	\label{fig:gazebo_sim}
\end{figure}

\subsection{Trajectory Optimization: ILQR}
\label{sec:trajopt}
This section describes the navigation of the robot using the nominal ILQR algorithm, the objective function design, and specific initializations for the robot. The goal is for the robot to move optimally from the center of the environment in \autoref{fig:gazebo_sim}a to within a $0.1m$ radius of the orange box attached to the right-hand corner of \autoref{fig:gazebo_sim}b. We separate the state and control cost terms for easier manipulation of cost components. For the state's instanteneous cost, we define a pseudo 	``smooth-abs" function,
\begin{align}
\small
\ell(\state_t) = \sqrt{\alpha + (\state_t - \state^\star)^T\, \textbf{diag}(\stateweight) \, (\state_t - \state^\star)},
\label{eq:state_cost}
\end{align}
where $\state^\star$ denotes the desired state, $\stateweight$ is a state penalty vector, and $\alpha$ is a constant that controls the curvature of the cost function: the lower the $\alpha$ value, the smoother is the robot's trajectory near the goal state. The $(\state - \state^\star)$ term encourages the robot to follow the nominal trajectory while driving toward the goal state. This $l_{12}$  function enforces reaching a desired target exactly in 3D space. Equation \ref{eq:state_cost} encourages the relative weighting of state-cost terms along different axes of motions. Inspired by \cite{Tassa12}, we choose a hyperbolic cosine control cost function (instead of a quadratic cost which gives disproportional controls in differrent portions of the state space) defined as:
\begin{align}
\ell(\control_t) = \alpha^2 \left(\text{cosh}(\textbf{\textit{w}}_u^T \control_t ) - 1\right),
\label{eq:control_cost}
\end{align}
where $\textbf{\textit{w}}_u$ is a control penalty vector. This cost function limits control outputs to an $\alpha$-sized neighborhood in $\textbf{\control}-$space. We set $\alpha$ to $10^{-4}$, $\textbf{\textit{w}}_u \text{ to } [1, 1, 1, 1]$ and $\textbf{\textit{w}}_x  \text{ to } [1, 1, 0.8]$. We found a time horizon, $T = 150$ to be appropriate for this task. We proceed as follows:
\begin{itemize}
	\item starting with an open-loop control schedule $\{\bar{\textbf{u}}_t\}$, (initialized to $[1.3, 0.8, 0.1]$), we generate the nominal states $\{\bar \state_t\}$
	\item replacing $\torque$ with $\control$ in \eqref{eq:inverse_dyn}, we compute the forward dynamics:
	\begin{align}
	\ddot{\textit{\state}} = - \massinertia^{-1} \coriolis \dot{\textit{\state}}  - \massinertia^{-1} \Bmat^T \left(\Smat \frictionvec - \frac{1}{\wheelrad} \control \right)
	\label{eq:inv_dyn}
	\end{align}
	\item we then obtain derivatives of $\textbf{\textit{f}}$ and those of $\ell$ from \eqref{eq:dyn_cost}
	\item starting at time $t = T-1$, we compute the associated nominal Q coefficients in \eqref{eq:Q_coeffs}, obtain the open and closed-loop gains, $\vectorgain_\control , \matrixgain_\control$,  and obtain the value function derivatives
	\item in the forward pass, we proceed from $t = \{0, \ldots, T-1\}$, and update the trajectories with a backtracking line search parameter, $0 < \varsigma \le 1$, as follows
	\begin{align}
	\hat \state(1)   = \state(1), \, 
	\hat \control(t) = \control(t) &+ \varsigma \, \vectorgain_\control(t) + \matrixgain_\control(t) (\hat \state(t) - \state(t))  \nonumber \\
	\hat \state(t+1) &= \textbf{\textit{f}}\left(\hat \state(t), \hat \control(t)\right)
	\end{align}
	\item the backward/forward pass informs of the change in cost $\Delta J^\star$, which is compared to the estimated reduction
	$
	\eta = \left(J(\control_{1, \ldots, T-1}) - J(\hat{\control}_{1, \ldots, T-1} \right) / \Delta(J(\rho)),
	$
	where
	\[
	\Delta(J(\rho)) = \rho \sum_{t=1}^{T-1} \vectorgain_{\control \tidx}^T \textbf{Q}_{\control\tidx} + \frac{\rho^2}{2} \sum_{i=1}^{T-1} \vectorgain_{\control\tidx}^T \textbf{Q}_{\control\control\tidx} \vectorgain_{\control\tidx}.
	\]
	\item we accept a trajectory only if $0 < c < \eta $ 
	(where we have chosen $c = 0.5$) following \cite{JacobsonMayne}'s recommendation.
	\item set $\nomstate_t = \state_t (t = 1, \ldots, T),\,\, \control_t = \bar{\control}_i$, reset $\rho$ and repeat the forward pass
	\item stop when $\Delta(J(\rho)) < \chi$ (where $\chi$ is a problem dependent term).
\end{itemize}

\subsection{Trajectory Optimization: IDG}
\label{sec:idg}
We initialized the adversarial inputs $\distbo$ from a Gaussian distribution $\sim \mathcal{N} (\bm{0}, \bm{2I})$ and augment the stage control cost of \eqref{eq:control_cost} as:
\begin{align}
\ell(\control_t, \distbo_t) = \alpha^2 \left(\text{cosh}(\actionweight^T \control_t)
- \gamma \,  \text{cosh}(\advactionweight^T \distbo_t)\right).
\label{eq:augmented_control_cost}
\end{align}
\noindent $\gamma \text{cosh}(\advactionweight^T \distbo_t)$ introduces a weighting term in the disturbing input, with $\gamma$ as the robustness parameter.
We set $\advactionweight$ to $[1, 1, 1, 1]$ and run the two player, zero-sum game erstwhile described.
%
%
Again, we compute the derivatives $\fcontrol$ and $\fdistbo$, calculate the associated derivatives in \eqref{eq:dyn_cost} and give the optimized iDG torque 
to the robot.
We initialized the nominal disturbance vectors $\nomdist$, and $\textit{\distbo}_\tidx$ as a multivariate Gaussian-filtered, random noise vectors $\sim \mathcal{N}(\textbf{0}, \textbf{2I})$. 
Our iDG process goes thus:
\begin{itemize}
	\item starting with an open-loop control and disturbance schedules $\{\bar{\textbf{u}}_t\}, \, \{\bar{\textbf{v}}_t\}$, we generate the nominal states $\{\bar \state_t\}$
	%
	%
	\item we then obtain the derivatives $\fcontrol$, $\fdistbo$, $\fcontroldistbo$, $\fcontrolcontrol$, $\fdistbodistbo$ and those of $\ell$ from \eqref{eq:dyn_cost}
	\item in a backward pass, we obtain the associated nominal Q coefficients, the open and closed-loop gains, 
	and obtain the improved value function coefficients with \eqref{eq:value_coeffs}
	\item in the forward pass, we proceed from $t = \{0, \ldots, T-1\}$, and update the trajectories with a backtracking linesearch parameter, $0 < \varsigma \le 1$, as follows
	\begin{align}
	\hat \state(1)   = \state(1), 
	\hat \control(t) = \control(t) &+ \varsigma \, \vectorgain_\control(t) + \matrixgain_\control(t) (\hat \state(t) - \state(t))  \nonumber \\
	\hat \distbo(t) = \distbo(t) + \varsigma \, \vectorgain_\distbo(t) &+ \matrixgain_\distbo(t) (\hat \state(t) - \state(t))  \nonumber \\
	\hat \state(t+1) &= \textbf{\textit{f}}\left(\hat \state(t), \hat \control(t), \hat \distbo(t)\right)
	\end{align}
	
	\item the backward/forward pass informs of the change in cost $\Delta J^\star$, which is compared to the estimated reduction
	$
	\eta = \frac{J(\control_{1, \ldots, T-1}, \distbo_{1, \ldots, T-1}) - J(\hat{\control}_{1, \ldots, T-1}, \hat{\distbo}_{1, \ldots, T-1})}{\Delta(J(\rho))},
	$
	where
	\begin{align}
	&\Delta(J(\rho)) = \rho \sum_{t=1}^{T-1}\left[ \vectorgain_\control(t)^T \textbf{Q}_\control (t)
	+ \vectorgain_\distbo(t)^T \textbf{Q}_\distbo (t) \right] + \ldots \nonumber \\
	& \ldots \frac{\rho^2}{2} \sum_{i=1}^{T-1}\left[ \vectorgain_\control(t)^T \textbf{Q}_{\control\control} (t) \vectorgain_\control(t) 
	+ \vectorgain_\distbo(t)^T \textbf{Q}_{\distbo\distbo} (t) \vectorgain_\distbo(t)
	\right] \nonumber
	\end{align}
	\item we accespt a trajectory only if $0 < c < \eta $ 
	(choosing $c = 0.5$)
	
	\item set $\state_t = \nomstate_{t, \, \{t = 1, \ldots, T\}},\, \control_t = \bar{\control}_\tidx, \,  \distbo_t = \bar{\distbo}_\tidx$, reset $\rho$ and repeat the forward pass
	
	\item stop when $\Delta(J(\rho)) < \chi$ (where $\chi$ is a problem dependent term).
\end{itemize}
\section{Results}  \label{sec:results}
We implement the policy sensitivity analysis using the guided policy search algorithm of \cite{Levine16}.
We then run the iDG optimization algorithm on the youbot and analyze the effectiveness of using our method against standard ILQG. The codes for reproducing the experiments can be found on github at \href{https://github.com/lakehanne/youbot/tree/rilqg/}{https://github.com/lakehanne/youbot/tree/rilqg}.

\subsection{Robustness Margins}
\begin{figure}[tb!]
    \centering
    \includegraphics[width=0.65\columnwidth,height=.45\columnwidth]{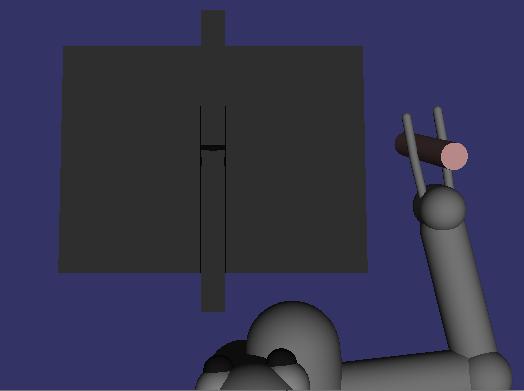}
    \caption{Policy Robustness Quantification Experiment}\label{fig:pol_quant}
\end{figure}

We first generate data used to train a global neural network policy by running our ILQG optimization scheme for the local controllers. We then run an adversarial local controller in closed-loop with the nominal agent's controller for various disturbance values,  $\gamma$, in a supervised learning of global neural network policies.
We task the PR2 robot (shown in \autoref{fig:pol_quant} and simulated in the MuJoCo physics simulator \cite{mujoco}), to insert a peg into the hole at the bottom of the slot. The model was learned with a mixture of $N=40$ Gaussians and we parameterized the local control laws with the deep neural network described in \cite{Levine16}. The difficulty of this task stems from the discontinuity in dynamics from the contact between the peg and the slot's surface.

We choose quadratic stage costs for both the state and the two controllers
\begin{align}  \label{eq:peg_robust_cost}
\augreward(\state_t, \textbf{\textit{u}}_t, \textbf{\textit{v}}_t) = \textbf{\textit{w}}_u \textbf{\textit{u}}_t^T \textbf{\textit{u}}_t + \textbf{\textit{w}}_x \| \textbf{\textit{d}}(\state_t) - \textbf{\textit{d}}^\star \|^2 - \gamma  \distbo_t^T \distbo_t 
\end{align}
where $\textbf{\textit{d}}(\state_t)$ denotes the end effector position at state $\state_t$ and $\textbf{\textit{d}}^\star$ denotes the desired end effector position at the base of the slot. The cost function quantifies the squared distance of the end effector to its desired position and energy of the the controller and adversary torque inputs. We set $\textbf{\textit{w}}_u$ and $\textbf{\textit{w}}_x$ to $10^{-6}$ and $1$, respectively.
\begin{figure}[tb!]
	\resizebox{\columnwidth}{!}{
		\begin{tikzpicture}
		\begin{loglogaxis}[
		ybar,
		anchor=north,
		/pgf/number format/fixed,
		/pgf/number format/precision=1,
		x tick label style={
			/pgf/number format/1000 sep=},
		title=\textbf{Optimal adversarial costs vs. \bm{$\gamma$}-penalty},
		ylabel=adversary cost, 
		xlabel=\bm{$\gamma$}-\textbf{penalty},
		enlargelimits=0.45,
		ybar,
		bar width=15pt,
		]
		\addplot
		coordinates {
			(1e-6, 2060998193.53)
		}; 
		\addplot
		coordinates {(1e-5, 1900342478.37)};
		\addplot
		coordinates {(1e-4, 1830342478.37)};
		\addplot
		coordinates {(1e-3, 1550342465.37)};
		\addplot
		coordinates {(1e-2, 1284488685.84)};
		\end{loglogaxis}
		\end{tikzpicture}
		\begin{tikzpicture}
		\begin{axis}[
		xmin=1.5,
		xmax=6,
		xtick={0.5,1.5,4.0,5.0,7.0},
		enlargelimits=false,
		nodes near coords align={vertical},
		title=\textbf{Optimal adversarial costs vs. \bm{$\gamma$}-penalty},
		xlabel=\bm{$\gamma$}-\textbf{penalty},
		enlargelimits=0.75,
		ybar,
		bar width=15pt,
		]
		\addplot
		coordinates {(0.5, -350.93)};
		\addplot
		coordinates {(1.5, -490.14)};
		\addplot
		coordinates {(2.0, -542.81)};
		\addplot
		coordinates {(3.0, -462.75)};
		\addplot
		coordinates {(5.0, 	-463.47000)};
		\addplot
		coordinates {(7.0, 	-260.89)};
		\end{axis}
		\end{tikzpicture}
	}
\caption{Robustness Analysis for Peg Insertion Task}
\label{fig:peg_insert}
\end{figure}
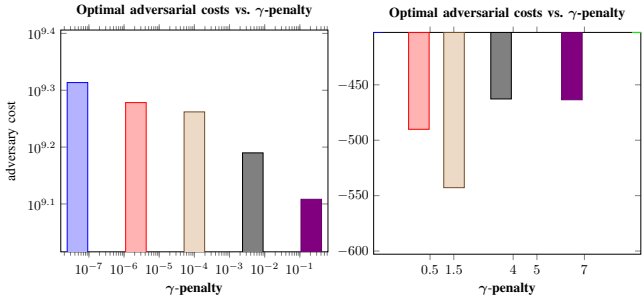
For various $\gamma$-values, we evaluate the robustness of the trained policy by training the adversary with the GPS algorithm, and observe its effect on the task performance. We run each experiment for $11$ GPS iterations.
\autoref{fig:peg_insert} shows that the adversary causes a sharp degradation in the controller performance for values of $\gamma < 1.5$. This corresponds to when the optimized policy is destabilized and the arm fails to reach the desired target.
\begin{table*}[tb!]
		\centering
		\begin{tabular}{cM{35mm}M{35mm}M{35mm}M{35mm}}
    \label{tbl:idg_results}
				$\state_1^\star$ & \includegraphics[height=8em,width=10em]{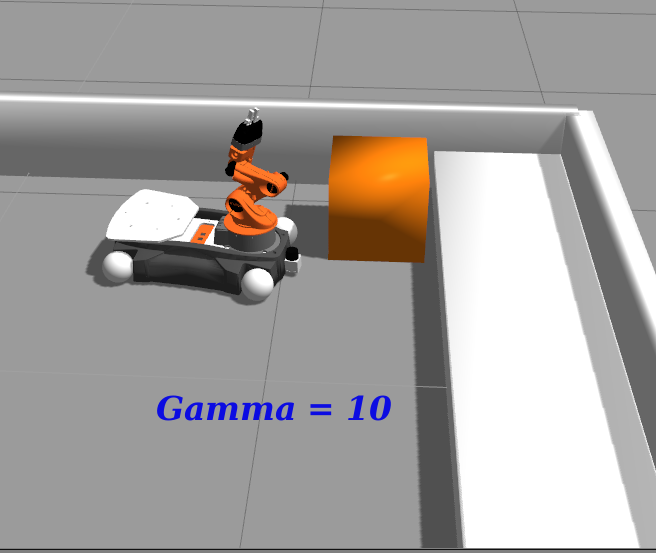} & \includegraphics[height=8em,width=10em]{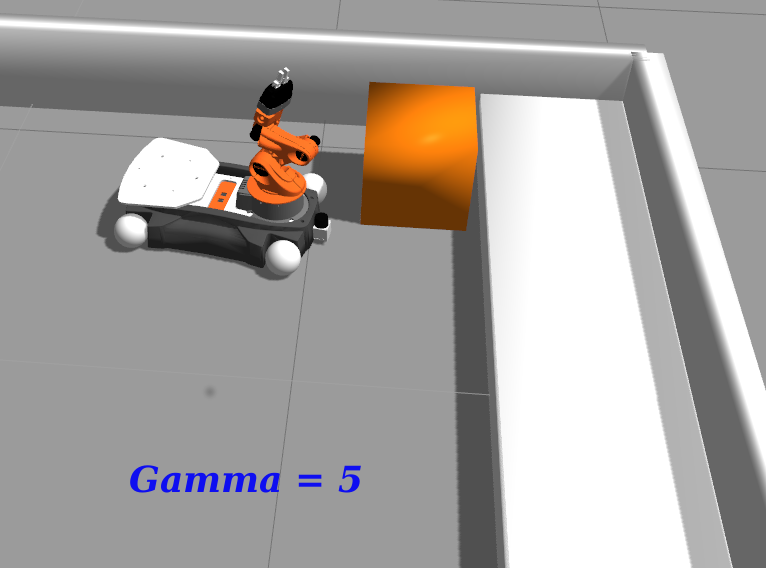} & \includegraphics[height=8em,width=10em]{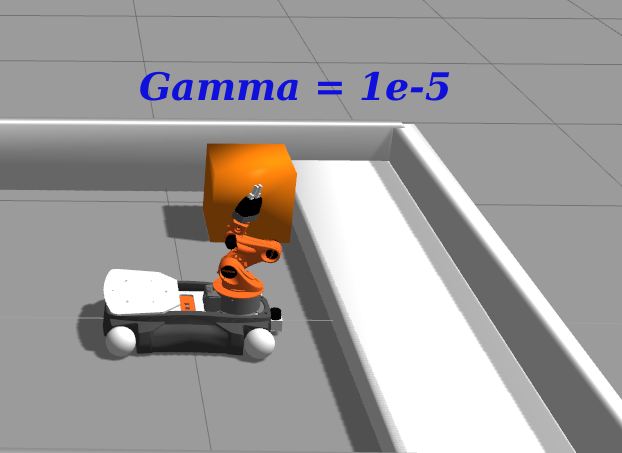} & \includegraphics[height=8em,width=10em]{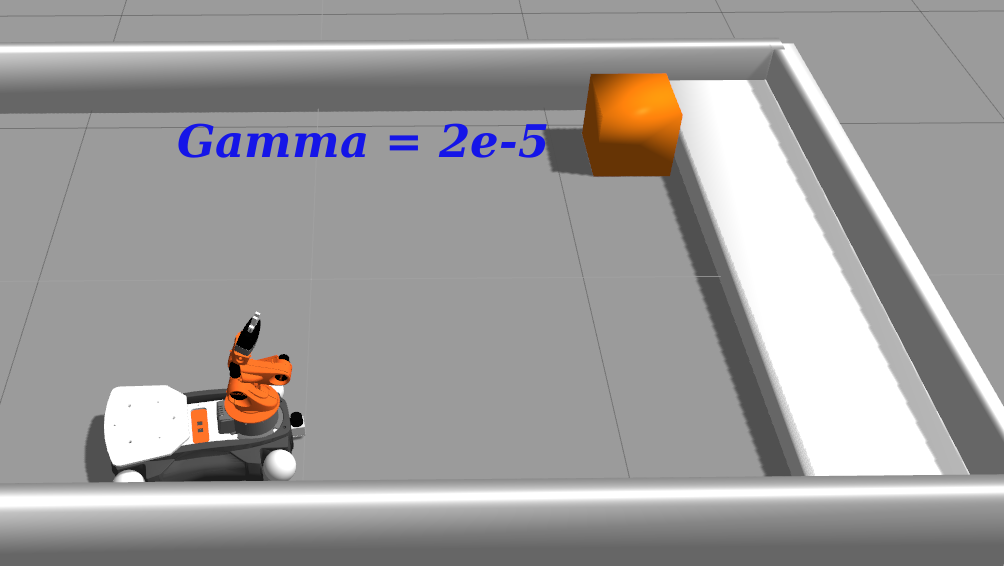} \\
				$\state^\star_2$ & \includegraphics[height=8em,width=10em]{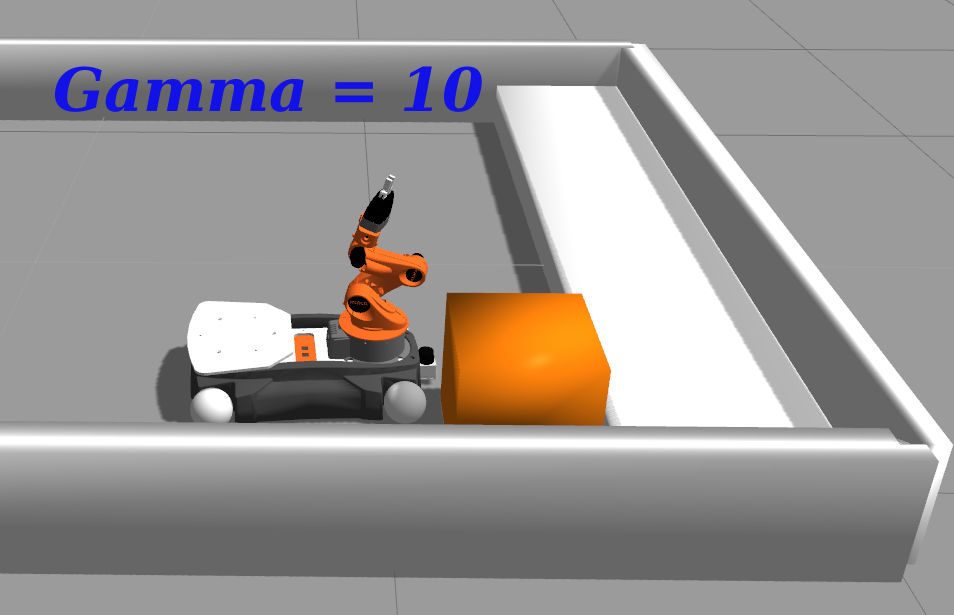} & \includegraphics[height=8em,width=10em]{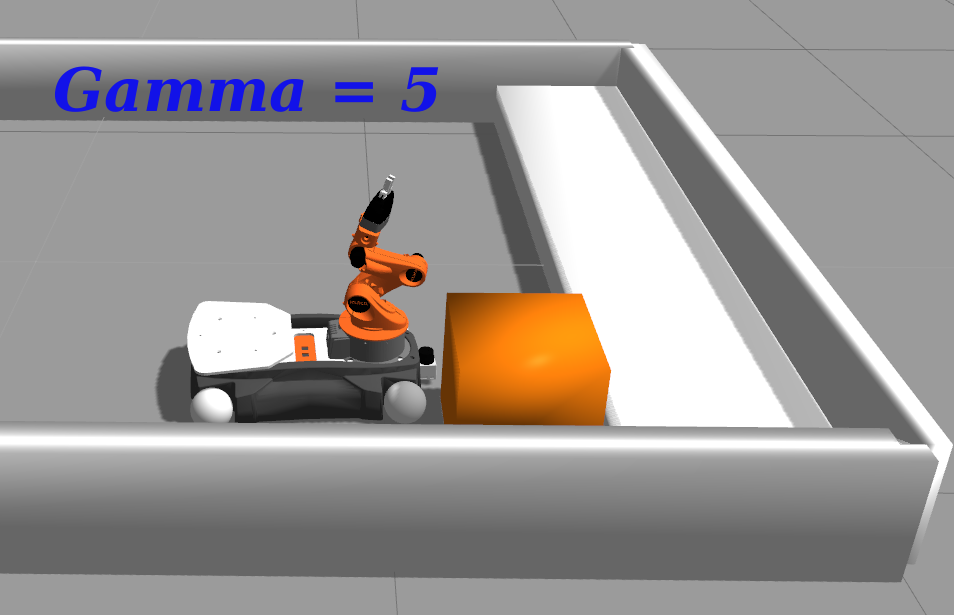} & \includegraphics[height=8em,width=10em]{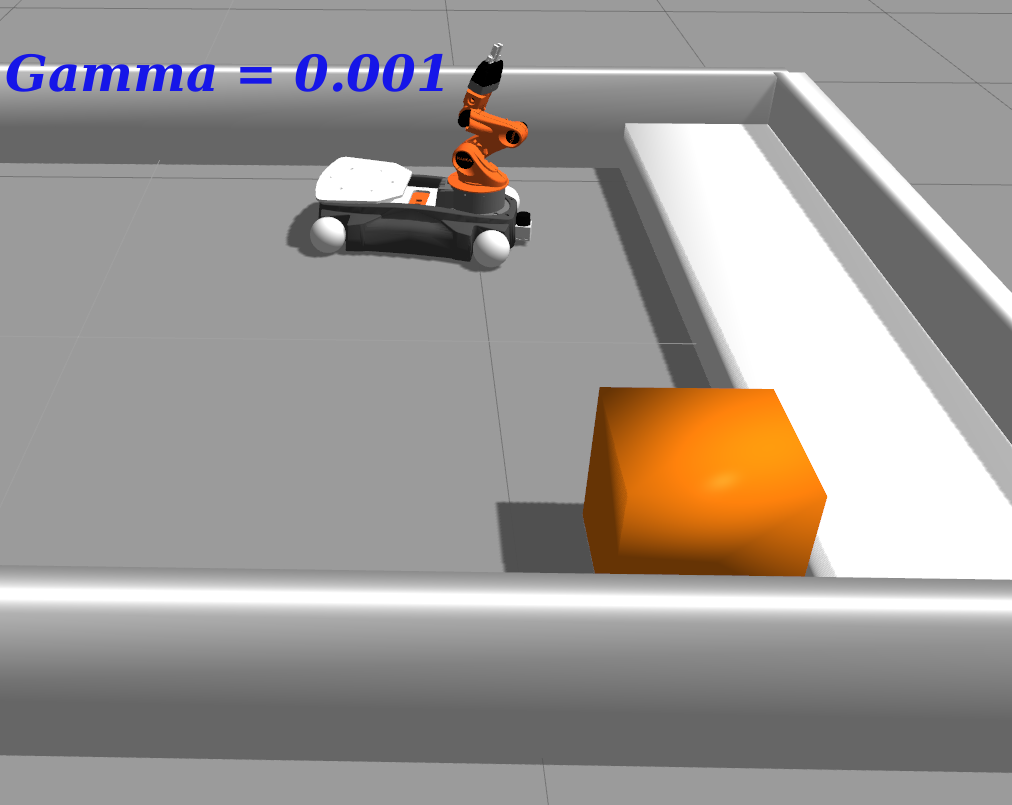} & \includegraphics[height=8em,width=10em]{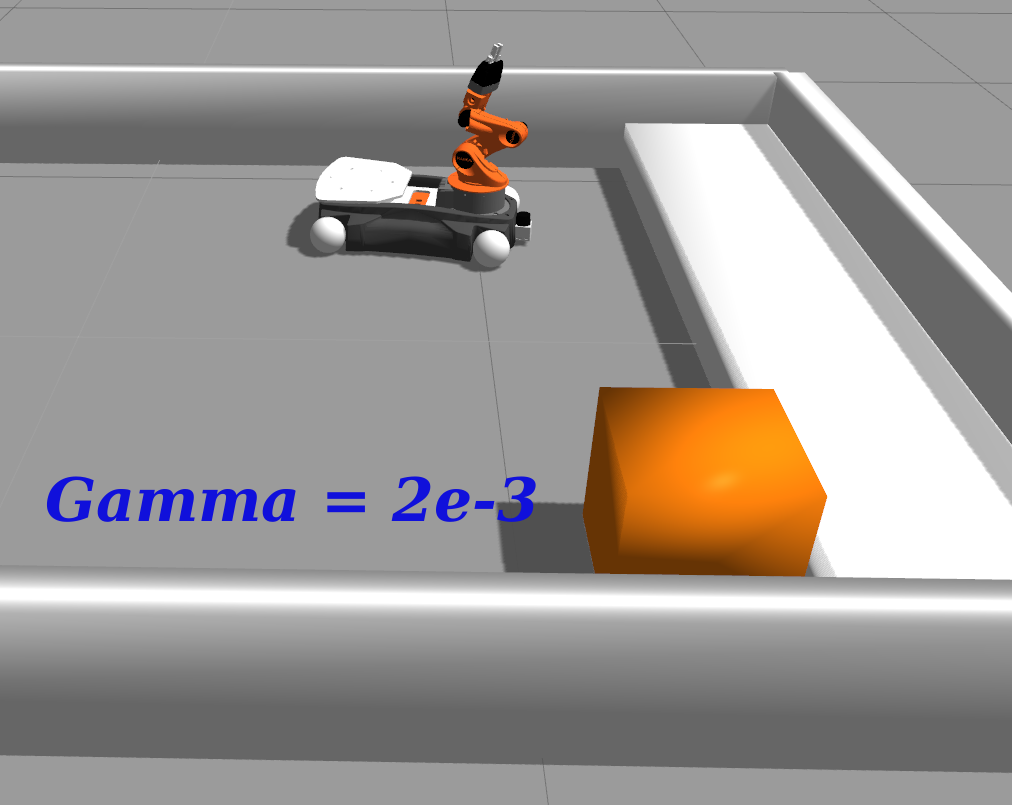} 
		\end{tabular}
		\caption*{State convergence of the KUKA robot with our iDG formulation given different goal states and varying $\gamma$-values}
\end{table*}
As values of $\gamma$ increase above 1.5, however, we observe that the adversary has a reduced effect on the given task.
Video of this result is on the robustness margins link on our website.

\subsection{ILQG-based Trajectory Optimization}
We evaluate the trajectory optimization algorithm on the youbot KUKA robotic platform. The iLQG result is best understood by watching the ILQG video result available here: \href{https://goo.gl/JhshTB}{https://goo.gl/JhshTB}.
In the video, observe that the arm on the robot vibrates with abrupt motions while navigating toward the desired goal. This is expected as the dynamics of the arm links were not included in the model of the platform. Regardless, the robot reaches the goal state after 76 seconds.

\subsection{Trajectory Optimization: iDG}
We found $T = 150$ to be a desirable stage horizon, initiate the disturbance from a multivariate Gaussian filtered noise $\mathcal{N}(0, 2)$, set the gains for the torques to $\{k_{F_x} = 10, k_{F_y} = 15, k_{F_\theta} = 1 \}$ respectively, and run the algorithm described in \ref{sec:idg} for various $\gamma$-disturbance values. We run two experiments: one with the goal at far left corner of the environment and the second with the goal state at the far right corner of the environment. We pose various adversarial inputs against the controller with values of $\gamma$ in the range $\{10, 5, 3, 1.5, 0.65, 0.1, 0.1, 10^{-5}, 2 \times10^{-3}, 2\times10^{-5}\}$ for both experiments.  The result showing the trajectory evolution and the position of the youbot after each iDG run for various $\gamma$ values is better appreciated by watching the videos available on our website: \hspace{3em} \href{https://goo.gl/JhshTB}{https://goo.gl/JhshTB}.

In both experiments I and II, observe that for values of $\gamma$ in the range $1.5 \le \gamma \le 10$, the robot drives smoothly and reaches the goal without the arm vibrating on the platform as in the ILQG video. This is despite that did not take the mass and inertia matrix components of the arm links into consideration.  When $\gamma \le 1.5$, we notice that the adversarial disturbance's effect on the overall system start getting pronounced. While the robot still reaches its desired goal state for $\gamma = 0.5 \text{ and } 1.5$, the motion of the arm is no longer stable. Indeed for critical values of $\gamma \le 10^{-3}$, the trajectory of the robot becomes perturbed as well as the balance of the robot on the arm. As $\gamma$ becomes very small (below $10^{-5}$), the robot's trajectory is undefined, and it converges to a spurious minimum as both experiments show. This $\gamma$ indices would correspond to the  $\mathcal{J}^\star_{\gamma^\star}$ that depicts unacceptable performance in \autoref{fig:tradeoff}.

There is a time-robustness tradeoff in solving a policy optimization scheme with our iDG algorithm/other nonlinear methods. While iLQG does achieve the goal in generally 3-4 iterations, iDG solves the same task in about 5-7 iterations. When speed is not crucial, and the safety of a real-world agent/its environment is important, iDG provides a viable alternative for designing safe policies. Without loss of generality, we envisage that our minimax iDG scenario is extensible to similar systems that compute gradients of a cost function or reward in achieving an optimal control task. Compared to Morimoto's work \cite{Morimoto03}, our minimax iDG does not have to learn the unmodeled disturbance with reinforcement learning after implementing the minimax player. We conjecture that \cite{Morimoto03}'s model lacked robustness due to the incorrect second order derivative of the value function recursions.
%
%
We thus identify the critical value of $\gamma$ to be between $0.1 - 10^{-3}$ as being the value that causes maximal disturbance in the robot's torque. To design a robust policy, one can consider $\gamma$ values in this range in order to design an smooth trajectory for this particular using the minimax iDG algorithm.

\section{Conclusions} \label{sec:conclusions}
Despite exhibiting near-optimal performance, high-dimensional policies often exhibit brittleness in the presence of adversarial agents, model mismatch or policy transfer \cite{kos17, Pinto17}. I
In this work, we have presented a way of identifying the greatest upper bound on a policy's sensitivity via a minimax framework. We have also presented an iDG framework that informs of how to make a policy robust to unmodeled disturbances, and uncertainities, up to a disturbance bound by converging to a saddle equilibrium. We ran four experiments in total to validate our  hypothesis and the videos of our results are available on our website. 

\providecommand\BIBentryALTinterwordstretchfactor{2.5}
\bibliographystyle{IEEEtran}
\bibliography{root}

\end{document}